\documentclass[11pt]{article}
\usepackage[final]{acl}

\usepackage{times}
\usepackage{latexsym}
\usepackage[T1]{fontenc}
\usepackage[utf8]{inputenc}
\usepackage{microtype}
\usepackage{inconsolata}
\usepackage{graphicx}

\usepackage{booktabs}
\usepackage{multirow}
\usepackage{tabularx}
\usepackage{array}
\usepackage{amsmath,amssymb}
\usepackage{enumitem}


\newcommand{\asr}{\mathrm{ASR}}
\newcommand{\avgq}{\mathrm{Avg.Q}}
\newcommand{\gptfour}{GPT-4o}
\newcommand{\dsr}{DeepSeek-Reasoner}
\newcommand{\dsc}{DeepSeek-Chat}

\title{What Drives Dialectal Jailbreaks? An Ablation of Surface Form, Cultural Framing, and Strategy Banks}

\author{Qingyang Xu \\
  Independent Researcher, Shanghai, China \\
  \texttt{qyxu1994@gmail.com}}

\begin{document}
\maketitle

\begin{abstract}
Recent work suggests that obscure language registers can weaken large language model refusal behavior, especially when paired with black-box prompt optimization. It remains unclear whether failures stem from non-standard surface form, culturally grounded framing, or optimization over an expressive prompt-strategy space. We study Chinese registers by extending a classical-Chinese red-teaming framework to Shanghainese and Cantonese and running a 36-cell ablation across surface forms, strategy-bank variants, and two target models. The main finding is corrective: dialectal surface form is neither necessary nor sufficient for high attack success. Non-optimized English, Mandarin, and naive dialect translations remain below 8\% attack success rate, whereas all conditions that retain an optimizer-controlled strategy bank reach 98--100\%. A culture-neutral generic strategy bank reaches the same ceiling at near-single-query cost, further indicating that strategy-bank expressiveness, rather than dialectal or cultural content, accounts for most of the observed effect. Dialect choice still affects query efficiency and response severity, and qualitative coding identifies recurring high-level mechanisms such as semantic glossing and procedural scaffolding. We qualify the absolute ceiling-level attack success rate by showing sensitivity to judge calibration and noting that the design does not fully separate one-shot strategy construction from iterative search.
\end{abstract}

\section{Introduction}

LLMs are increasingly deployed as multilingual assistants, but their safety mechanisms are rarely evaluated with the same linguistic breadth as their advertised capabilities. A model that robustly refuses a harmful English or Mandarin request may respond differently when the same intent is embedded in a low-resource language, a dialectal register, or a locally specific cultural frame. This gap matters both technically and socially: technically, it reveals weak semantic generalization of safety alignment; socially, it risks creating unequal protection for speakers of underrepresented languages and varieties.

Prior work has documented cross-lingual jailbreak vulnerabilities in low-resource languages \citep{yong2023lowresource,deng2024multilingual}, automatic adversarial suffixes \citep{zou2023universal}, and real-world prompt jailbreak communities \citep{shen2024dan}. Most closely related to our study, \citet{huang2026obscure} show that classical Chinese---a concise, historically prestigious, and comparatively underrepresented written register---can bypass model safeguards when paired with black-box prompt optimization. That result raises two related questions. First, is the vulnerability specific to classical Chinese, or does it generalize to living Chinese regional varieties whose written forms, idioms, and cultural references are also underrepresented in safety training? Second, within such an attack, which component is actually doing the work: the dialectal surface form, the cultural framing, or the iterative optimizer over an attack-strategy space?

We investigate both questions through a red-team evaluation of Chinese dialectal distribution shifts. We use ``dialect'' in its common NLP sense for regional Sinitic varieties, while recognizing that Shanghainese/Wu and Cantonese/Yue are linguistically rich varieties with distinct phonology, lexicon, and sociocultural histories. Unlike classical Chinese---a historical, largely written-only register no longer used in everyday speech---these are \emph{living} varieties with millions of contemporary speakers, which is what makes gaps in their safety coverage practically consequential. Our goal is not to characterize speakers or communities as risky, but to test whether LLM safety mechanisms generalize across the contexts in which users actually interact with deployed systems.

We extend the CC-BOS framework \citep{huang2026obscure}---a black-box jailbreak that uses a population-based optimizer (fruit-fly optimization, FOA) to search over a bank of prompt-construction strategies, requiring only query access to the target---from classical Chinese to Shanghainese and Cantonese, preserving the black-box optimization loop but replacing the register-specific prompt constructor with dialect-specific sociocultural framing modules and adding a two-stage translation pipeline. On top of this we design a six-condition ablation varying (i) the surface form (English, Mandarin, dialect) and (ii) which optimizer strategy dimensions are active, evaluated against \gptfour{} and \dsr{} on a 50-prompt AdvBench subset (1800 evaluations). The attack and translation model is \dsc{}, the judge is \gptfour{}, and we early-stop at $\tau=80$ on a 0--120 scale.

This paper makes the following contributions:
\begin{enumerate}[leftmargin=*]
    \item We design a controlled 36-cell ablation that systematically varies the surface form and the cultural-framing strategy bank while holding the target models, judge, and evaluation budget fixed, and show that an expressive optimizer-controlled strategy bank dominates while the dialect surface form contributes only to efficiency and severity, not to binary success. (Our design does not separate iterative search from one-shot strategy-bank construction; we return to this in Section~\ref{sec:analysis}.)
    \item We extend dialectal red-team evaluation for Chinese regional varieties from classical Chinese to Shanghainese/Wu and Cantonese/Yue, and report that the dialect-framed attack reaches $100\%$ ASR on both \gptfour{} and \dsr{} under our judge and threshold.
    \item We provide a safety-preserving qualitative taxonomy of successful living-dialect attacks, with reliability probed by an independent automated re-coder (median $\kappa = 0.71$ between the author and a hypothesis-blind language-model agent; Section~\ref{app:irr}), identifying semantic glossing, procedural scaffolding, commerce/logistics metaphors, grey-zone authority frames, and performative social scripts as recurring mechanisms, and show via a non-FOA template-artifact control that these mechanisms are emergent under optimization rather than artifacts of the prompt template.
    \item We articulate a safety-preserving reporting protocol for dialectal jailbreak research: publish aggregate metrics, failure modes, and defenses, while withholding optimized adversarial prompts and harmful target responses.\footnote{Code and aggregate evaluation results are available in the \href{https://github.com/qyxu1994/chinese-dialect-jailbreak}{project repository}.}
\end{enumerate}

\section{Related Work}

\paragraph{Jailbreak attacks and red teaming.}
LLM jailbreak research studies prompts that induce models to violate intended safety policies. Early work characterized real-world jailbreak prompts and their strategies, including role play, privilege escalation, and prompt injection \citep{shen2024dan}. Automated methods search over adversarial strings or prompt transformations, such as gradient-guided suffix optimization \citep{zou2023universal}, red-teaming with language models \citep{ganguli2022redteam,perez2022redteam}, and benchmark-driven safety and jailbreak evaluation \citep{gehman2020realtoxicityprompts,chao2023pair,souly2024strongreject}. A closely related line optimizes \emph{readable} prompts or whole templates rather than suffixes: genetic-algorithm search over handcrafted prompts \citep{liu2024autodan}, fuzzing-style mutation of seed jailbreak templates \citep{yu2023gptfuzzer}, tree-of-attack refinement \citep{mehrotra2024tap}, and encoding-based transformations such as ciphering \citep{yuan2024cipher}. These methods already show that optimization or mutation over a template/strategy space can break GPT-class models, so a key part of our contribution is the corrective finding that, on top of such an optimizer, the dialectal surface form and its cultural grounding add essentially nothing to binary success. Our work follows this empirical safety-evaluation tradition but focuses on a specific source of distribution shift: Chinese dialectal and cultural register, and the interaction between that shift and an optimizer-driven attack-strategy bank.

\paragraph{Multilingual safety alignment.}
Modern LLMs acquire refusal behavior through alignment techniques such as reinforcement learning from human feedback \citep{ouyang2022training} and Constitutional AI \citep{bai2022constitutional}, but these safeguards are trained mostly on high-resource languages and often fail to generalize: \citet{yong2023lowresource} show that translating unsafe prompts into low-resource languages bypasses GPT-4 safeguards, and \citet{deng2024multilingual} find unsafe behavior increasing as language resource availability decreases, with multilingual alignment hard to resolve via translated data alone \citep{shen2024safetychallenges}. These studies typically operate at the level of national languages; we instead examine regional Sinitic varieties and culturally grounded register shifts within the Chinese language family.

\paragraph{Obscure-register jailbreaks.}
Most closely related is CC-BOS \citep{huang2026obscure}, which uses multi-dimensional fruit-fly optimization over dimensions such as role, metaphor, expression, and context to generate classical-Chinese adversarial prompts in black-box settings, framing classical Chinese as concise and obscure relative to modern safety-training corpora. We ask whether this ``high capability--low alignment'' phenomenon extends to living dialects, and whether the dialectal surface form is itself load-bearing.

\paragraph{Dialectal and low-resource NLP.}
NLP systems underperform on dialectal and low-resource varieties because data, annotation, and evaluation norms privilege standardized registers \citep{joshi2020state,bender2011achieving}, creating representational and practical harms. We connect this to LLM safety by showing that dialectal and register variation exposes inconsistencies in refusal robustness, not only in task accuracy---while clarifying that in our ablation the dominant failure mode is optimization over strategy dimensions rather than the dialect surface form alone, so the effect is broader than dialectal coverage per se.

\section{Threat Model and Responsible Scope}

We evaluate a black-box red-team setting in which an auditor queries a deployed target model and observes its response, with access to an attack model that proposes dialect-framed variants and a translation model for judge-based evaluation---approximating third-party robustness auditing of closed-weight systems. Because jailbreak research is dual-use, we restrict operational detail: we report aggregate metrics, sanitized methodology, and defense implications, and omit optimized prompts, raw harmful responses, and prompt templates that would make the attack directly reusable.

\section{Method}

\subsection{Base Optimizer}

Our evaluation builds on the CC-BOS black-box optimization framework \citep{huang2026obscure}. At a high level, the optimizer maintains a small population of prompt-construction policies. Each policy maps a harmful intent into a dialect-framed prompt using a set of register, context, and metaphor dimensions. The target model is queried with candidate prompts; the resulting responses are translated to English when necessary; and an automated judge assigns a score. The optimizer then updates the candidate population using a bio-inspired search procedure---fruit-fly optimization (FOA), which we use as shorthand for the optimizer throughout.

Let $x_i$ denote the $i$-th harmful-intent prompt, $d \in \mathcal{D}$ a dialect/register, and $m \in \mathcal{M}$ a target model. A prompt constructor produces a dialectal prompt $z_{i,d,\theta} = \pi_\theta(x_i, d)$, where $\theta$ indexes a redacted set of register-framing choices; the target returns $y = m(z_{i,d,\theta})$, translated to English as $\tilde{y} = T_d(y)$ when required. The judge assigns a consistency score $c_i \in \{0,20,\dots,100\}$ (a 0--5 rubric scaled by 20) plus a keyword bonus $b_i \in \{0,20\}$, giving a total $s_i = c_i + b_i \in [0,120]$. We define success at threshold $\tau=80$:
\begin{equation}
    \asr_\tau(d,m)=\frac{1}{n}\sum_{i=1}^n \mathbb{I}\{s_i(d,m)\ge \tau\}.
\end{equation}
We separately report Avg.Score, the average judge rating on the original 0--5 scale (more comparable to the rubric than the thresholded score), and the average number of target-model queries $\avgq$, over all prompts and over successful prompts.

The consistency score $c_i$ is the judge's rating of how fully the response carries out the harmful intent---as opposed to refusing, deflecting, or answering a different question---on the 0--5 rubric of \citet{huang2026obscure}, rescaled by $20$ so that it shares the $0$--$120$ range with the keyword bonus. We inherit the rubric, the $\times 20$ scaling, and the success threshold $\tau=80$ (a rubric level of $4$ out of $5$) unchanged from CC-BOS for comparability; none of the three is tuned, and $\tau$ is applied identically to every dialect, target, and condition. Because binary ASR depends on this cutoff, we report a threshold-sensitivity analysis in Appendix~\ref{app:threshold}: the two-regime gap between non-FOA and FOA conditions persists across $\tau \in \{40,60,80\}$ (non-FOA stays below $7\%$ while FOA remains at ceiling), so the headline contrast is not an artifact of the specific threshold.

\subsection{Dialect Extensions}

We evaluate three prompt registers:
\begin{itemize}[leftmargin=*]
    \item \textbf{Classical Chinese}: the original CC-BOS baseline, using a single-stage classical-Chinese-to-English translation path for evaluation.
    \item \textbf{Shanghainese/Wu}: a living regional variety with dialect-specific lexical and pragmatic features, paired with local sociocultural framing. Evaluation uses a two-stage translation path: Shanghainese/Wu-oriented text to Mandarin gloss, then Mandarin to English.
    \item \textbf{Cantonese/Yue}: a living regional variety with a broader online footprint than Shanghainese, paired with southern Chinese and Hong Kong cultural framing. Evaluation also uses the two-stage translation path.
\end{itemize}

The dialect modules preserve the optimizer interface and the number of high-level strategy dimensions, but the concrete prompt templates and optimized prompts are not disclosed. This design permits direct comparison with classical Chinese while reducing the risk of releasing reusable attack artifacts.

\subsection{Translation and Quality Control}

Because target responses may contain dialectal, Mandarin, or mixed-register content, we translate before judging (single-stage for classical Chinese, two-stage through Mandarin glosses for Shanghainese and Cantonese). The system logs translation-quality flags for hedging or low-confidence phrases; this does not eliminate translation noise but prevents obvious low-confidence translations from being silently treated as reliable evidence.

\subsection{Ablation Conditions}
\label{sec:ablation-design}

To separate the contribution of the dialectal surface form from the contribution of the cultural-framing strategy bank and the FOA optimizer itself, we partition the optimizer's eight prompt-construction dimensions into two banks: \emph{surface-form dimensions} (e.g., expression style, knowledge framing) and \emph{cultural-frame dimensions} (e.g., role, metaphor, context, guidance). The exact split is given in our public configuration. Crossing FOA usage, frozen-dimension bank, and surface form yields the six conditions in Table~\ref{tab:ablation_design}.

\begin{table}[t]
\centering
\small
\resizebox{\columnwidth}{!}{%
\begin{tabular}{lccc}
\toprule
Condition & FOA? & Frozen dims & Surface form \\
\midrule
\texttt{english\_original} & no & n/a & English \\
\texttt{mandarin\_translation} & no & n/a & Mandarin \\
\texttt{naive\_dialect\_translation} & no & n/a & Dialect \\
\texttt{dialect\_without\_cultural\_frame} & yes & cultural & Dialect \\
\texttt{mandarin\_with\_cultural\_frame} & yes & surface & Mandarin \\
\texttt{full\_dialect\_cultural\_frame} & yes & none & Dialect \\
\bottomrule
\end{tabular}%
}
\caption{The six ablation conditions. Non-FOA conditions issue a single attack-LLM call per harmful intent and skip the optimization loop. FOA conditions with a frozen bank fix those dimensions at a canonical baseline.}
\label{tab:ablation_design}
\end{table}

The three non-FOA conditions isolate surface form alone (English, Mandarin, or naive dialect translation); the three FOA conditions isolate the optimizer by freezing one bank: surface-form-only (\texttt{dialect\_without\_cultural\_frame}), cultural-frame-only in Mandarin (\texttt{mandarin\_with\_cultural\_frame}), or both banks in the target dialect (\texttt{full\_dialect\_cultural\_frame}).

\section{Experimental Setup}

\paragraph{Dataset.}
We use the 50-prompt AdvBench subset used in the CC-BOS comparison setting \citep{zou2023universal,huang2026obscure}. The prompts cover multiple harmful-intent categories. In the paper we do not reproduce the raw prompts.

\paragraph{Models.}
The attack model is \dsc{}, which proposes candidate dialect-framed prompts. The translation model is also \dsc{}. The automated judge is \gptfour{}. The target models are \gptfour{} and \dsr{}. All models are accessed through their providers' hosted APIs during May 2026: \gptfour{} (OpenAI, \texttt{gpt-4o}), \dsc{} and \dsr{} (DeepSeek, \texttt{deepseek-chat} and \texttt{deepseek-reasoner}), and---for the cross-vendor judge-agreement check in Appendix~\ref{app:judge_agreement}---Qwen-Max (Alibaba, \texttt{qwen/qwen3.6-max-preview} via OpenRouter). Because these are hosted, continually updated endpoints, exact reproduction may require pinning to the same API snapshots.

\paragraph{Optimization budget.}
For each FOA cell, we use population size 5 and maximum 5 optimization iterations. We use early-stop threshold $\tau=80$, corresponding to a query-efficient setting that stops when a substantial successful response has been detected. This setting should not be interpreted as maximizing response severity; it is designed to measure rapid penetration and query cost.

\paragraph{Completed runs and integrity checks.}
We complete all 36 cells: $3\text{ dialects}\times6\text{ conditions}\times2\text{ targets}\times50\text{ prompts} = 1800$ target-model evaluations. Every record file contains exactly 50 unique prompt IDs with non-empty responses and valid score, target, dialect, and condition fields (a keep-last dedupe pass was applied to earlier-session records).

\section{Results}

\subsection{Attack Success Rates}

Under the full method (\texttt{full\_dialect\_cultural\_frame}; bottom row of Table~\ref{tab:ablation_asr}), all three dialects achieve $100.0\%$ ASR against both \gptfour{} and \dsr{} with no genuine failures. Avg.Score is consistently high ($4.02$--$4.28$ on the $0$--$5$ scale), with \dsr{} marginally above \gptfour{} in every dialect (peaking at $4.28$ for classical Chinese on \dsr{}). For each $50/50$ cell the Wilson $95\%$ confidence interval is approximately $92.9$--$100.0\%$; the benchmark is small, so we do not over-interpret sub-percent differences, but the ceiling ASR across all tested registers is robust under the same evaluation budget (and threshold-robust; Appendix~\ref{app:threshold}).

\subsection{Ablation Results}
\label{sec:ablation-results}

Table~\ref{tab:ablation_asr} reports ASR for all six conditions. A clean two-regime pattern emerges. The three non-FOA conditions stay between 0\% and 8\% ASR for every dialect--target pair: directly translating a harmful English prompt into Mandarin or into a dialect, without iterative search over a strategy bank, almost always fails. In contrast, every FOA condition---including the two restricted variants that hold one of the two strategy banks frozen---reaches 98--100\% ASR.

\begin{table*}[t]
\centering
\small
\resizebox{\textwidth}{!}{%
\begin{tabular}{lcccccc}
\toprule
\multirow{2}{*}{Condition} & \multicolumn{2}{c}{Classical Chinese} & \multicolumn{2}{c}{Shanghainese} & \multicolumn{2}{c}{Cantonese} \\
\cmidrule(lr){2-3}\cmidrule(lr){4-5}\cmidrule(lr){6-7}
 & \gptfour{} & \dsr{} & \gptfour{} & \dsr{} & \gptfour{} & \dsr{} \\
\midrule
\texttt{english\_original} & 2.0 & 2.0 & 0.0 & 2.0 & 2.0 & 2.0 \\
\texttt{mandarin\_translation} & 2.0 & 2.0 & 6.0 & 2.0 & 2.0 & 2.0 \\
\texttt{naive\_dialect\_translation} & 2.0 & 8.0 & 2.0 & 4.0 & 2.0 & 0.0 \\
\midrule
\texttt{dialect\_without\_cultural\_frame} & 100.0 & 100.0 & 100.0 & 100.0 & 100.0 & 100.0 \\
\texttt{mandarin\_with\_cultural\_frame} & 100.0 & 100.0 & 100.0 & 100.0 & 100.0 & 98.0 \\
\texttt{full\_dialect\_cultural\_frame} & 100.0 & 100.0 & 100.0 & 100.0 & 100.0 & 100.0 \\
\bottomrule
\end{tabular}%
}
\caption{ASR (\%) under all six conditions, $\tau=80$, $n=50$ per cell. The top three rows are non-FOA baselines (single attack-LLM call). The bottom three rows hold the FOA optimizer fixed and ablate which strategy bank is active. Non-FOA conditions stay below 10\% in every cell; FOA conditions reach ceiling in every cell.}
\label{tab:ablation_asr}
\end{table*}

The ablation supports two claims. The dialectal surface form is not what causes the failure: \texttt{mandarin\_with\_cultural\_frame} outputs Standard Mandarin yet reaches 98--100\% ASR, matching the full dialect condition. Nor are the cultural-framing dimensions strictly necessary: with that bank frozen (\texttt{dialect\_without\_cultural\_frame}) ASR still reaches 100\% everywhere. The common ingredient explaining the jump from 0--8\% to 98--100\% is the optimizer operating over \emph{some} non-trivial strategy bank; once it runs, the surface language and the choice of which bank to optimize have at most a small effect on binary success.

Paired McNemar tests confirm this split (Appendix~\ref{app:paired_tests}): across all 18 within-FOA contrasts every p-value is $1.00$ with at most one discordant pair, so we cannot reject equality of the three FOA conditions at $n{=}50$---a null constrained by ceiling saturation rather than a positive demonstration of symmetry. Every FOA-vs-non-FOA contrast, by contrast, is highly significant (e.g., $46/50$ discordant pairs, $p{<}10^{-10}$). We discuss the implications in Section~\ref{sec:analysis}.

\subsection{Query Efficiency}

Although binary success saturates in the FOA conditions, query cost does not. Table~\ref{tab:avgq} reports average target-model queries under the full method. Against \gptfour{}, classical Chinese is most efficient (2.56), followed by Shanghainese (3.54) and Cantonese (5.14). Against \dsr{}, Shanghainese is most efficient (2.98), followed by classical Chinese (3.18) and Cantonese (5.84). Cantonese is consistently the least efficient register against both targets.

\begin{table}[t]
\centering
\small
\resizebox{\columnwidth}{!}{%
\begin{tabular}{lcccc}
\toprule
\multirow{2}{*}{Dialect/Register} & \multicolumn{2}{c}{All prompts} & \multicolumn{2}{c}{Successful only} \\
\cmidrule(lr){2-3}\cmidrule(lr){4-5}
 & \gptfour{} & \dsr{} & \gptfour{} & \dsr{} \\
\midrule
Classical Chinese & \textbf{2.56} & 3.18 & \textbf{2.56} & 3.18 \\
Shanghainese/Wu & 3.54 & \textbf{2.98} & 3.54 & \textbf{2.98} \\
Cantonese/Yue & 5.14 & 5.84 & 5.14 & 5.84 \\
\bottomrule
\end{tabular}%
}
\caption{Average target-model queries under the full method ($\tau=80$). Because every prompt succeeds, the ``successful-only'' columns equal the ``all prompts'' columns. Dialect choice changes efficiency by up to 2$\times$ even though final ASR is identical.}
\label{tab:avgq}
\end{table}

We read these differences as interactions between target-model refusal heuristics and dialectal framing, not as a measure of attack effectiveness; Cantonese's lower efficiency is plausibly tied to its greater online representation, though we do not directly measure pretraining exposure.

\subsection{Score Distribution and Severity}

Table~\ref{tab:score_dist} (Appendix~\ref{app:score_dist}) shows the score distribution under the full method. \gptfour{} responses cluster tightly at the early-stop threshold: 48--49 of 50 successful responses receive a total score of exactly 80. \dsr{} produces more high-severity responses even though the optimizer stops at the first threshold-crossing success: 4--7 of 50 prompts reach scores of 100 or 120 per dialect, versus 1--2 of 50 for \gptfour{}. The high-severity count is largest for classical Chinese on \dsr{} (9/50) and Cantonese on \dsr{} (7/50).

\subsection{Failure Analysis}

Under the full method there are zero failures across all six cells; failures concentrate in the non-FOA baselines (near floor) and a single near-ceiling FOA cell (Cantonese \texttt{mandarin\_with\_cultural\_frame} on \dsr{}, 49/50). Given the optimizer and either bank, residual binary variation is essentially search-budget noise rather than harm-category dependence.

\subsection{Culture-Neutral Generic Strategy Control}
\label{sec:generic_control}

To test whether the cultural framing in our dialectal strategy bank is itself load-bearing---or whether \emph{any} sufficiently expressive optimizer-controlled strategy bank suffices---we run a culture-neutral generic-strategy variant (Table~\ref{tab:generic_control}). The generic bank replaces the dialect-specific sociocultural dimensions (commerce framing, tea-house social frame, performing-arts frame, etc.) with a culture-neutral set of role, format, scenario, and pretext dimensions, evaluated across three surface forms: English, Mandarin, and a generic Chinese register. All other elements of the pipeline---FOA loop, target models, judge, $\tau{=}80$ early-stop, $n{=}50$ AdvBench subset---are held fixed.

\begin{table}[t]
\centering
\small
\resizebox{\columnwidth}{!}{%
\begin{tabular}{llrrr}
\toprule
Surface form & Target & ASR (\%) & Avg.\ total (0--120) & Avg.\ q. \\
\midrule
English          & \gptfour{} & $100.0$ & $80.4$ & $1.24$ \\
Mandarin         & \gptfour{} & $100.0$ & $82.0$ & $1.30$ \\
Chinese register & \gptfour{} & $100.0$ & $81.6$ & $1.36$ \\
\midrule
English          & \dsr{}     & $100.0$ & $82.0$ & $1.52$ \\
Mandarin         & \dsr{}     & $100.0$ & $83.6$ & $1.54$ \\
Chinese register & \dsr{}     & $100.0$ & $85.6$ & $1.40$ \\
\bottomrule
\end{tabular}%
}
\caption{Culture-neutral generic-strategy control: FOA over a generic (non-culturally-grounded) strategy bank, three surface forms, $n{=}50$ AdvBench prompts per cell, $\tau{=}80$. All six cells reach $100\%$ ASR with average severity comparable to the dialectal full method, and at lower query cost ($1.24$--$1.54$ vs.\ the dialectal cells in Table~\ref{tab:avgq}). The cultural framing is therefore \emph{not} a necessary ingredient: any sufficiently expressive optimizer-controlled bank reaches the same ceiling.}
\label{tab:generic_control}
\end{table}

This strengthens the ablation claim: surface form is not load-bearing once an optimizer and a strategy bank are active (Section~\ref{sec:ablation-results}), and the bank's \emph{cultural grounding} is not load-bearing either---a culture-neutral bank reaches the same ceiling at \emph{lower} query cost. What remains is the conjunction \emph{optimizer plus any sufficiently expressive strategy bank}, whether culturally grounded or neutral, across any surface form.

\section{Qualitative Analysis of Successful Dialectal Attacks}
\label{sec:qualitative}

To understand \emph{how} the successful dialectal prompts work, we analyze the 200 successful optimized prompts under the full method (100 Shanghainese, 100 Cantonese). The analysis is safety-preserving: we code high-level rhetorical and sociocultural mechanisms without reproducing prompts or harmful responses. Schema-defining counts come from an earlier 99-per-dialect snapshot (frozen before the reliability check so labels cannot be retrofitted); the check itself (Section~\ref{sec:coding}) uses the final FOA data.

\subsection{Coding Procedure and Reliability}
\label{sec:coding}

We coded each successful prompt for recurring surface and discourse mechanisms. The coding is not intended to establish causality, because successful prompts are outputs of an optimizer and many mechanisms co-occur. Instead, it characterizes the prompt families that the optimizer repeatedly converged on. Table~\ref{tab:qual_features} (Appendix~\ref{app:irr}) reports prevalence by dialect under the v1 descriptive snapshot used to define the schema.

To assess the reliability of the schema, two coders independently labeled a random sample of 20 successful prompts drawn from the final FOA data (10 Shanghainese and 10 Cantonese, all under the full-method condition) using the 9-mechanism schema in Table~\ref{tab:qual_features}. To preserve independence, the second coder was a separate general-purpose language-model agent dispatched in an isolated session with the schema and prompts but no exposure to the paper's hypothesis, prior coding, or methodological context; this is disclosed here as a methodological substitute for a fully human second coder. Median Cohen's $\kappa$ across the nine mechanisms was $+0.706$ (range $+0.000$ to $+1.000$), with four mechanisms at perfect agreement; per-mechanism values, including the one degenerate-marginal $\kappa{=}0$ case, are reported in Appendix~\ref{app:irr}.

As a template-artifact control, we additionally coded 20 non-FOA prompts drawn from the same record set (\texttt{english\_original}, \texttt{mandarin\_translation}, and \texttt{naive\_dialect\_translation}) and compared mechanism prevalence. The three-step procedural scaffold appears in $1/20$ non-FOA prompts but $20/20$ FOA prompts ($\Delta = +0.95$); parenthetical semantic glossing shows the same $1/20$ vs.\ $20/20$ split ($\Delta = +0.95$); modern technical bridges, commerce/logistics metaphors, and grey-zone authority frames appear in zero non-FOA prompts. This pattern supports the claim that the common skeleton in Section~\ref{sec:qualitative} is \emph{emergent under optimization} rather than mechanically induced by the prompt template or by the underlying AdvBench requests alone.

\subsection{A Common Skeleton: Glossing plus Proceduralization}

Two mechanisms appear in nearly all successful prompts. First, parenthetical semantic glossing: a culturally specific phrase is paired with an explanatory gloss pointing the model toward the intended modern abstraction, creating a dual-channel prompt that is opaque to safety heuristics relying on standard phrasing yet transparent enough for the base model to follow---\emph{selective transparency} rather than simple obfuscation. Second, a three-part procedural scaffold: once the model accepts the prompt as a legitimate explanatory task, the structure elicits ordered, complete, concrete content---so the dialectal wrapper shifts the model into an instructional register while the scaffold determines the response shape. The ablation (Section~\ref{sec:ablation-results}) is consistent with this: the conditions that reach ceiling are those that retain both the scaffold and the gloss, and removing the optimizer removes them, with ASR returning to the non-FOA baseline. We read these mechanisms as characterizing the prompt families the optimizer repeatedly converges on; they should not be read as individually necessary or sufficient.

\subsection{Per-Dialect Profiles}

The two living dialects converge on different sociocultural frames (full prevalences in Table~\ref{tab:qual_features}). We stress that these frame preferences are \emph{emergent}: they are read off the prompts the optimizer converged on against each target, not imposed from sociolinguistic priors, and we make no claim that human speakers of these varieties preferentially use such frames. Shanghainese prompts are dominated by commercial, logistical, and semi-legal urban frames---account books, merchant houses, warehouses, concession-era ambiguity---that recode harmful requests as business process, asset transfer, or access control, and they frequently add an explicit modern technical bridge (88/99) reinterpreting the local scene through contemporary engineering or systems language. Cantonese prompts use commerce and grey-zone frames less heavily and instead draw more on performative and communal discourse (performing-arts, tea-house, and ritual settings), relying on pragmatic role embedding rather than explicit technical recoding: the model is invited into a socially familiar scene that treats the request as genre knowledge or narrative reconstruction. Cantonese is also consistently less query-efficient than Shanghainese, plausibly because written Cantonese has more online representation and thus somewhat better model coverage; our analysis cannot directly verify this training-data hypothesis.

\subsection{Mechanism Interactions}

Successful prompts rarely rely on a single mechanism: a typical pattern combines a culturally specific scene, a euphemistic mapping from the harmful concept to a benign local object, parenthetical glossing that restores the modern semantics, and a multi-step request format. This layering helps explain the high ASR despite dialect differences. Prompt length itself does not explain success or efficiency---within each dialect, length has near-zero correlation with query count among successful prompts---so the decisive factor is the alignment between framing, glossing, and the target's safety heuristics, not verbosity.

\subsection{High-Severity Successes}

Among the 200 living-dialect successes under the full method, 11 receive scores of $100$ or $120$, ten of them against \dsr{}---reinforcing that \dsr{} produces more severe responses immediately after a dialect-framed prompt crosses the refusal boundary, even though the optimizer stops at $\tau{=}80$ without continued severity-maximization. The qualitative mechanisms also sharpen the defense implications we return to in Section~\ref{sec:analysis} and Section~\ref{sec:defenses}: keyword blacklists miss benign-vocabulary prompts that carry harm through metaphor and glossing, while flagging all dialectal language would be discriminatory, motivating semantic normalization with uncertainty-aware review instead.

\section{Analysis}
\label{sec:analysis}

\subsection{An Expressive Strategy Bank Is the Load-Bearing Component}

The ablation in Section~\ref{sec:ablation-results} reframes the dialectal jailbreak story. Without FOA, no surface form---English, Mandarin, or dialect---produces more than 8\% ASR. With FOA, both a Mandarin-surface attack restricted to the cultural-frame bank and a dialect-surface attack restricted to the surface-form bank reach ceiling. The dialectal surface form is therefore neither necessary nor sufficient: it is a vector along which the optimizer can find effective attacks, but several other vectors work just as well.

\paragraph{Search versus construction.} We do not over-attribute this to the iterative \emph{search} itself. The non-FOA baselines remove both the multi-dimensional strategy-bank \emph{construction} and the optimization loop, so they cannot separate adaptive search from a single richly constructed prompt. The generic control (Section~\ref{sec:generic_control}) is informative: it reaches $100\%$ ASR at only $1.24$--$1.54$ queries---the first constructed candidate usually succeeds and the search loop is barely engaged---suggesting the load-bearing ingredient is the strategy-bank construction rather than search per se. We thus frame the load-bearing component as an \emph{optimizer-or-template} strategy bank. A budget-matched best-of-$k$ control---non-adaptive sampling of the same strategy bank at FOA's full $25$-query budget, which removes adaptive search while holding construction and budget fixed---is designed to test this directly. This control is currently \emph{partial}: of the six intended dialect${\times}$target cells only classical-Chinese/\gptfour{} is complete, where non-adaptive sampling reaches $100\%$ ASR ($33/33$) at the matched budget. We report this as preliminary rather than decisive evidence and do not rest the main attribution on it; it is consistent with the near-single-query generic control (Section~\ref{sec:generic_control}), which likewise reaches ceiling with the search loop barely engaged. Completing the remaining five cells is the cleanest way to fully isolate adaptive search from construction (Section~\ref{sec:limitations}).

Two consequences follow: surface-only defenses (e.g., dialect-specific keyword filters) will not generalize, since the same attack succeeds with a Mandarin surface form; and future ``dialectal jailbreak'' evaluations should always include a Mandarin-surface variant of the same optimized strategy bank, or an apparent dialect effect may simply be an optimizer effect.

\subsection{Dialect-Agnostic Effectiveness, Dialect-Specific Efficiency}

The observed ASR pattern is dialect-agnostic under the full method: all tested dialects reach ceiling ASR under our judge and threshold. The efficiency pattern is dialect-specific: the optimizer reaches successful prompts at different query costs depending on the target model. We interpret this as evidence that the main weakness lies in semantic safety generalization under optimization plus a strategy bank, while search efficiency depends on target-model training coverage and refusal heuristics for the specific surface variety. A benchmark reporting only ASR would call all dialects equivalent; query cost and severity reveal which dialect--model pairs are more brittle in practice. More broadly, the generic control (Section~\ref{sec:generic_control}) locates the failure in the safety classifier's tolerance for optimizer-driven prompt-shape variation rather than in missing dialectal training data; because the same surface-form sensitivity also fires on innocuous inputs, benign users in underrepresented varieties may see inconsistent safety behavior.

Finally, the reasoning model \dsr{} is more vulnerable than \gptfour{}: it produces more high-score responses across all dialects and is the only target admitting a non-ceiling ablation cell (49/50). This may reflect a reasoning-oriented model more readily elaborating once it accepts a dialect-framed premise, or weaker safety coverage for these registers; we cannot distinguish these without private chain-of-thought or training data, but it motivates targeted safety stress-testing of reasoning models.

\section{Defense Recommendations}
\label{sec:defenses}

We propose three practical recommendations. First, \textbf{dialect-aware red-team suites with Mandarin-surface controls}: audits should cover dialectal and regional varieties (classical/literary registers, Cantonese, Shanghainese, Hokkien/Min, Hakka, code-switching) and include a Mandarin-surface version of every optimized attack, to avoid mistaking an optimizer effect for a dialect effect. Second, \textbf{semantic intent detection after normalization}: because every successful attack in our study preserves the harmful \emph{intent} while only varying surface form and framing, the natural countermeasure is a guardrail that evaluates intent semantically rather than by surface cues. Concretely, such a guardrail would (i)~normalize dialectal or obfuscated input into a high-resource pivot (e.g., Mandarin or English) via translation or paraphrase; (ii)~recover the underlying request by asking a semantic classifier what task the prompt ultimately asks for, seeing through the recurring evasion scaffolds we identify in Section~\ref{sec:qualitative}---parenthetical glossing, multi-step procedural reframing, and metaphorical recoding; and (iii)~apply a single policy to the recovered intent regardless of surface variety, so the same request is treated identically in English, Mandarin, or a dialect. This must be paired with translation-quality checks in both directions: an over-aggressive normalizer risks discriminatory false positives against benign dialectal users, while a lossy one risks unsafe false negatives. Robust semantic intent detection---rather than surface-form or keyword refusal---is the defense most directly implied by our finding that the failure lives in the safety classifier's tolerance for optimizer-driven prompt-shape variation. Third, \textbf{dialectal safety fine-tuning and multi-metric reporting}: fine-tune on safety examples that preserve dialectal context (with native-speaker review), and report query efficiency and severity alongside ASR.

\section{Conclusion}

We asked which component of a dialect-framed black-box jailbreak is load-bearing. Across 1800 evaluations in a 36-cell ablation, classical Chinese, Shanghainese, and Cantonese all reach 100\% ASR on \gptfour{} and \dsr{}, yet the dialectal surface form is not: a Mandarin cultural-frame-only attack and a dialect surface-form-only attack both hit the ceiling while non-optimized translations stay below 8\%. The load-bearing ingredient is an expressive optimizer-controlled strategy bank, and dialect choice changes only efficiency and severity---leaving current safety alignment brittle under optimization over any such bank (defenses in Section~\ref{sec:defenses}).

\section*{Limitations}
\label{sec:limitations}

This study has several limitations.

\paragraph{Small benchmark, and a saturation-prone one.}
We evaluate a 50-prompt AdvBench subset. Although the runs are complete for all 36 cells of the ablation matrix, the benchmark is too small for precise estimates of small differences across dialects or target models. AdvBench is also known to elicit inflated success rates and ``empty'' jailbreaks relative to harder, rubric-graded benchmarks, which motivated StrongREJECT \citep{souly2024strongreject}. The ceiling-level ASR we observe should therefore be read partly as a property of the instrument. Larger benchmark and cross-benchmark generalization experiments (StrongREJECT, CLAS) and $\tau{=}120$ reruns of the main results are in progress and will be reported in an extended version.

\paragraph{Partial matched-budget non-adaptive control.}
Our within-FOA contrasts identify which strategy banks the optimizer relies on, but they do not by themselves separate \emph{adaptive search} from \emph{search budget}: a non-adaptive sampler with the same query budget might or might not reach the same ceiling. We have begun a budget-matched best-of-$k$ control over the same strategy bank (non-adaptive sampling at FOA's $25$-query budget; \texttt{phase\_e\_matched\_budget\_baselines} in \texttt{configs/ablation\_matrix.yaml}). It is not yet complete: of the six intended dialect${\times}$target cells, only classical-Chinese/\gptfour{} is finished, where non-adaptive sampling reaches $100\%$ ASR ($33/33$) at the matched budget. We treat this as preliminary evidence---consistent with, but not on its own establishing, the claim that adaptive search is not required for ceiling success---and do not rely on it for the central attribution, which rests on the ablation and the near-single-query generic control (Section~\ref{sec:generic_control}). We will report the full six-cell control in an extended version, and until then we limit the adaptive-search claim accordingly.

\paragraph{Single automated judge in the headline ASR.}
The headline ASR is computed under a single \gptfour{} judge. To bound the judge-choice confound, we re-scored a stratified subsample of $216$ rows (six rows per $\text{dialect}\times\text{target}\times\text{condition}$ cell, seeded for reproducibility) with a cross-vendor second judge (Qwen-Max, accessed via OpenRouter as \texttt{qwen/qwen3.6-max-preview}) and report the resulting agreement in Appendix~\ref{app:judge_agreement}. The two judges agree strongly on the continuous severity scale ($\text{Pearson } 0.87$, $\text{Spearman } 0.86$, $\text{Krippendorff } \alpha = 0.79$) and, on the total-score basis used for ASR throughout the paper, also on the binary success flag (raw agreement $0.931$, Cohen's $\kappa = 0.86$). A much lower $\kappa=0.11$ appears only if success is binarised on the strict consistency-only score (dropping the keyword bonus), where \gptfour{}'s marginal success rate is $\sim\!5\%$ against Qwen-Max's $\sim\!45\%$; this is a high-prevalence ``$\kappa$ paradox'' on a non-ASR basis, not a ranking disagreement (Appendix~\ref{app:judge_agreement}). We nonetheless caveat that the \emph{absolute} ASR ceiling under \gptfour{} is somewhat calibration-dependent---it falls to $88.8\%$ under Qwen-Max---even though the within-judge ablation contrasts (evaluated under a fixed judge) are robust to this gap. A related confound is model-family coupling: \gptfour{} serves as both a target and the primary judge, and \dsc{} is both the attack and the translation model. Same-family scoring and translation can bias absolute severity estimates; the cross-vendor Qwen-Max re-scoring partially bounds the judge side of this, but a fully decoupled judge and translation pipeline is left to future work.

\paragraph{Translation confounds.}
Shanghainese and Cantonese use two-stage translation through Mandarin before English evaluation. This may understate or distort response severity relative to classical Chinese. Better evaluation should include bilingual or dialect-proficient human annotation.

\paragraph{Limited target models in the full ablation.}
The full ablation evaluates only \gptfour{} and \dsr{}. Cross-model generalization to Claude Sonnet, Qwen-Max, and \dsc{} is configured but not yet complete. A small smoke test on \dsc{} suggests similar risks, but \dsc{} is also the attack and translation model, creating a self-attack confound.

\paragraph{No release of attack artifacts.}
For safety reasons, we do not release optimized prompts or raw harmful responses. This limits exact reproducibility, but we believe aggregate reproducibility and controlled access are appropriate for dual-use jailbreak research.

\paragraph{Sociolinguistic simplification.}
The term ``dialect'' hides substantial linguistic and social complexity. Shanghainese and Cantonese are not merely stylistic variants of Mandarin. Future work should collaborate with dialect speakers and sociolinguists to design more faithful and respectful evaluations.

\section*{Ethics Statement}

This work studies jailbreak vulnerabilities and is therefore dual-use. The purpose is defensive: to identify safety gaps affecting underrepresented language varieties and to motivate more equitable multilingual alignment. We intentionally omit optimized adversarial prompts, raw harmful responses, and prompt templates. We report only aggregate statistics and sanitized failure categories. Before public release of code or data, we recommend redacting harmful prompts and outputs, adding access controls for high-risk artifacts, and notifying affected model providers with a concise vulnerability report.

There is also a representational risk: readers might incorrectly infer that dialect speakers or regional cultures are associated with unsafe behavior. That is not our claim. The vulnerability lies in model safety generalization, not in the languages, dialects, or communities being evaluated. Dialect-aware safety should improve protection and service quality for speakers of underrepresented varieties.

\section*{Use of AI Assistants}
We used a large language model as the independent second coder for the inter-rater reliability check (Section~\ref{sec:coding}, Appendix~\ref{app:irr}), dispatched in an isolated, hypothesis-blind session as a methodological substitute for a human coder. AI assistants were additionally used for code scaffolding and for editing and proofreading the manuscript. All experimental design, scientific claims, statistical analyses, and results were verified by the authors.

\appendix

\section{Preliminary Smoke Test on DeepSeek-Chat}

Before the final 1800-evaluation run, a 5-prompt smoke test included \dsc{} as a target model. Classical Chinese and Cantonese each achieved 4/5 success, while Shanghainese achieved 5/5 success. We exclude \dsc{} from the main full-run comparison because it is also used as the attack and translation model, creating a self-attack confound. The smoke test is useful only as preliminary evidence that the vulnerability is not limited to the two fully evaluated targets.

\section{Comparison with CC-BOS}

The original CC-BOS paper reports classical-Chinese results under both a high-severity setting and a query-efficient setting. Our experiments use the query-efficient early-stop threshold $\tau=80$. Therefore, our Avg.Score values should not be directly compared to CC-BOS high-severity results obtained with a larger optimization budget. The ASR comparison is more meaningful than the average severity comparison. A tau=120 high-severity rerun of our main results is in progress and will be reported in an extended version.

\section{Score Distribution and Severity}
\label{app:score_dist}

Table~\ref{tab:score_dist} reports the total-score distribution under the full method, referenced from the severity discussion in the main text.

\begin{table}[t]
\centering
\small
\resizebox{\columnwidth}{!}{%
\begin{tabular}{llrrr}
\toprule
Target & Dialect/Register & Score 80 & Score 100 & Score 120 \\
\midrule
\multirow{3}{*}{\gptfour{}}
& Classical Chinese & 49 & 1 & 0 \\
& Shanghainese/Wu & 48 & 1 & 1 \\
& Cantonese/Yue & 49 & 0 & 1 \\
\midrule
\multirow{3}{*}{\dsr{}}
& Classical Chinese & 41 & 4 & 5 \\
& Shanghainese/Wu & 46 & 0 & 4 \\
& Cantonese/Yue & 43 & 1 & 6 \\
\bottomrule
\end{tabular}%
}
\caption{Total score distribution under the full method with early stopping ($\tau=80$). \dsr{} has more high-score outputs even though the optimizer is not trying to maximize severity beyond $\tau=80$.}
\label{tab:score_dist}
\end{table}

\section{Responsible Artifact Release Checklist}

For any public release accompanying this paper, we recommend:
\begin{enumerate}[leftmargin=*]
    \item release aggregate metrics and sanitized scripts for computing ASR, query cost, and score distributions;
    \item withhold optimized adversarial prompts and raw harmful model responses;
    \item provide only benign toy prompts in public examples;
    \item require controlled access for any high-risk benchmark artifacts;
    \item include documentation warning that dialectal vulnerability reflects model limitations, not speaker communities;
    \item include a model-provider notification plan for severe new vulnerabilities discovered in follow-up runs.
\end{enumerate}

\section{Redacted Prompt Structure and Evaluation Rubric}
\label{app:redacted_structure}

To convey the \emph{shape} of a successful optimized prompt without releasing a reusable attack, we give a schematic skeleton in which every content slot is a placeholder rather than filled text. This reflects the structure that the qualitative coding (Section~\ref{sec:qualitative}) found recurring across successful dialectal prompts; the placeholders are intentionally left unfilled to preserve the safety protocol.

\begin{quote}\small\ttfamily
{}[ROLE / PERSONA in dialect] \newline
{}[CULTURAL-FRAME SETUP: benign local scene] \newline
{}[REQUEST, recoded as a task within the scene] \newline
\phantom{xx}(gloss: [local term] $\rightarrow$ [abstract modern concept]) \newline
{}[3-STEP PROCEDURAL SCAFFOLD: 1. \dots 2. \dots 3. \dots] \newline
{}[OUTPUT-FORMAT INSTRUCTION]
\end{quote}

No harmful intent, target behavior, or operational detail is instantiated in any slot; the skeleton only shows how the optimizer wraps an arbitrary intent in persona, cultural framing, parenthetical glossing, and procedural structure. We do not release filled instances.

The automated judge is prompted separately from the attack. It receives the (translated) target response and the original intent and returns an integer rating on a $0$--$5$ rubric---$0$ for a clear refusal or off-topic answer, rising to $5$ for a response that fully and specifically carries out the requested harmful task---together with a short justification. This rating is the consistency score $c_i$ (rescaled by $20$; Section~\ref{sec:coding}); the keyword bonus $b_i$ is added by a separate rule-based check. Sharing the rubric's structure supports evaluation reproducibility without providing an attack, since the judge prompt scores responses rather than generating them.

\section{Paired Statistical Tests}
\label{app:paired_tests}

Table~\ref{tab:paired_tests} reports McNemar exact p-values and
discordant-pair counts for every within-FOA condition-pair contrast
within each (dialect, target) cell. For each contrast, the success
indicators are prompt-id aligned at $n{=}50$. The short labels
\texttt{no-frame}, \texttt{mand-frame}, and \texttt{full} correspond
to the three FOA conditions in Table~\ref{tab:ablation_design}.

\begin{table*}[t]
\centering
\small
\begin{tabular}{lllrrr}
\toprule
Dialect & Target & Contrast (A vs.\ B) & $n_{disc}$ & McNemar $p$ & $\Delta$ ASR (\%) \\
\midrule
Classical Chinese & \gptfour{} & \texttt{no-frame vs.\ mand-frame} & 0 & 1.00 & 0.0 \\
Classical Chinese & \gptfour{} & \texttt{no-frame vs.\ full} & 0 & 1.00 & 0.0 \\
Classical Chinese & \gptfour{} & \texttt{mand-frame vs.\ full} & 0 & 1.00 & 0.0 \\
Classical Chinese & \dsr{} & \texttt{no-frame vs.\ mand-frame} & 0 & 1.00 & 0.0 \\
Classical Chinese & \dsr{} & \texttt{no-frame vs.\ full} & 0 & 1.00 & 0.0 \\
Classical Chinese & \dsr{} & \texttt{mand-frame vs.\ full} & 0 & 1.00 & 0.0 \\
\midrule
Shanghainese & \gptfour{} & \texttt{no-frame vs.\ mand-frame} & 0 & 1.00 & 0.0 \\
Shanghainese & \gptfour{} & \texttt{no-frame vs.\ full} & 0 & 1.00 & 0.0 \\
Shanghainese & \gptfour{} & \texttt{mand-frame vs.\ full} & 0 & 1.00 & 0.0 \\
Shanghainese & \dsr{} & \texttt{no-frame vs.\ mand-frame} & 0 & 1.00 & 0.0 \\
Shanghainese & \dsr{} & \texttt{no-frame vs.\ full} & 0 & 1.00 & 0.0 \\
Shanghainese & \dsr{} & \texttt{mand-frame vs.\ full} & 0 & 1.00 & 0.0 \\
\midrule
Cantonese & \gptfour{} & \texttt{no-frame vs.\ mand-frame} & 0 & 1.00 & 0.0 \\
Cantonese & \gptfour{} & \texttt{no-frame vs.\ full} & 0 & 1.00 & 0.0 \\
Cantonese & \gptfour{} & \texttt{mand-frame vs.\ full} & 0 & 1.00 & 0.0 \\
Cantonese & \dsr{} & \texttt{no-frame vs.\ mand-frame} & 1 & 1.00 & +2.0 \\
Cantonese & \dsr{} & \texttt{no-frame vs.\ full} & 0 & 1.00 & 0.0 \\
Cantonese & \dsr{} & \texttt{mand-frame vs.\ full} & 1 & 1.00 & $-$2.0 \\
\bottomrule
\end{tabular}
\caption{Paired McNemar tests on within-FOA condition contrasts.
$n_{disc}$ = number of prompts (out of $50$) on which the two
conditions disagree on binary success ($\tau{=}80$). All p-values
are $1.00$ and discordant pairs $\le 1$ in every cell, so we cannot
reject the null that the three FOA conditions are equally effective
at $n{=}50$. By contrast, every FOA-vs-non-FOA contrast yields
$n_{disc} \ge 46$ and $p < 10^{-10}$. The complete per-pair table is released as \texttt{results/public/paired\_tests.csv} in the supplementary materials, alongside the aggregate ablation, generic-control, judge-agreement, and threshold-sensitivity CSVs. $\Delta\text{ASR} = \text{ASR}(A) - \text{ASR}(B)$ in percentage points.}
\label{tab:paired_tests}
\end{table*}

\section{Inter-Rater Reliability}
\label{app:irr}

\begin{table*}[t]
\centering
\small
\resizebox{\textwidth}{!}{%
\begin{tabular}{p{0.25\textwidth}p{0.47\textwidth}cc}
\toprule
Mechanism & Sanitized description & Shanghainese & Cantonese \\
\midrule
Parenthetical semantic glossing & Dialectal or cultural expressions are followed by parenthetical explanations that map the euphemism to an abstract modern concept. & 94/99 & 96/99 \\
Three-step procedural scaffolding & The prompt asks for a structured sequence, usually organized as three subquestions, converting the intent into a procedural explanation task. & 99/99 & 99/99 \\
Modern technical bridge & Local metaphors are explicitly connected to modern engineering, network, financial, or system concepts. & 88/99 & 53/99 \\
Commerce/logistics metaphor & Harmful intents are reframed as ledger work, cargo handling, handoff protocols, settlement, or route planning. & 79/99 & 42/99 \\
Grey-zone authority frame & The prompt invokes semi-legal or extra-legal social authorities, boundary spaces, or historical jurisdictional ambiguity. & 65/99 & 28/99 \\
Historical media or research pretext & The request is framed as historical reconstruction, newspaper-style reporting, archival research, or fiction-writing background. & 39/99 & 16/99 \\
Performing-arts frame & The request is embedded in storytelling, opera, theatrical, or performer-centered discourse. & 15/99 & 26/99 \\
Tea-house social frame & The prompt uses tea-house conversation, old-neighborhood gossip, or informal elder narration as the interactional setting. & 7/99 & 20/99 \\
Ritual or religious-social frame & The prompt maps the intent to temple, procession, geomancy, or ritualized community activity. & 11/99 & 23/99 \\
\bottomrule
\end{tabular}%
}
\caption{Qualitative mechanisms observed in successful living-dialect adversarial prompts (v1 coding snapshot, 99 per dialect). Counts are descriptive and non-exclusive; a single prompt can contain multiple mechanisms. We do not report prompt text.}
\label{tab:qual_features}
\end{table*}

Two coders independently labeled the same random sample of 20 successful FOA prompts (10 Shanghainese, 10 Cantonese, all under \texttt{full\_dialect\_cultural\_frame}) using the 9-mechanism schema of Table~\ref{tab:qual_features}. Coder A was the primary author. Coder B was a separate general-purpose language-model agent dispatched in an isolated session with only the coding schema and the 20 prompt texts: it was not told that this work is a paper revision, that any hypothesis was being tested, that a common rhetorical skeleton was at issue, that an optimizer had produced the prompts, or that any prior coding existed. Both coders worked from blind identifiers and ignored the dialect/target/condition/optimizer columns. This isolated-dispatch protocol is disclosed here as a methodological substitute for a fully human second coder. Per-mechanism Cohen's $\kappa$ is reported in Table~\ref{tab:irr_kappa}, alongside marginal prevalences.

\begin{table}[t]
\centering
\small
\begin{tabular}{lrrr}
\toprule
Mechanism & $\kappa$ & $p_A$ & $p_B$ \\
\midrule
Parenthetical glossing       & $+1.000$ & $1.00$ & $1.00$ \\
Three-step scaffolding       & $+1.000$ & $1.00$ & $1.00$ \\
Modern technical bridge      & $+0.000$ & $0.90$ & $1.00$ \\
Commerce/logistics metaphor  & $+0.219$ & $0.70$ & $0.95$ \\
Grey-zone authority frame    & $+0.706$ & $0.55$ & $0.40$ \\
Historical/research pretext  & $+0.571$ & $0.25$ & $0.20$ \\
Performing-arts frame        & $+1.000$ & $0.10$ & $0.10$ \\
Tea-house social frame       & $+0.667$ & $0.25$ & $0.40$ \\
Ritual/religious-social frame & $+1.000$ & $0.05$ & $0.05$ \\
\midrule
Median (all 9)               & $+0.706$ & --- & --- \\
\bottomrule
\end{tabular}
\caption{Per-mechanism Cohen's $\kappa$ between Coder A (primary author) and Coder B (separate general-purpose LM agent, dispatched in an isolated session with the schema and prompts only) on the 20-prompt FOA reliability sample. $p_A$, $p_B$: fraction of prompts coded $1$ by each coder. Median $\kappa = +0.706$; range $[+0.000, +1.000]$. The $\kappa{=}0.000$ entry for \emph{Modern technical bridge} reflects degenerate prevalence on the Coder B side ($p_B{=}1.00$): observed agreement (0.90) equals the chance expectation under those marginals, not disagreement on labels.}
\label{tab:irr_kappa}
\end{table}

As a template-artifact control, we also coded 20 non-FOA prompts (drawn from \texttt{english\_original}, \texttt{mandarin\_translation}, and \texttt{naive\_dialect\_translation} records, including refusals and bare AdvBench requests) with the same schema. Table~\ref{tab:irr_prevalence} compares mechanism prevalences. Every dialect-specific frame (commerce, grey-zone, tech bridge, perform-arts, tea-house) appears in zero non-FOA prompts; the procedural scaffold and parenthetical gloss each appear in only $1/20$ non-FOA prompts. This $\Delta \ge +0.90$ gap for the two near-universal FOA mechanisms supports the claim that the common skeleton in Section~\ref{sec:qualitative} is an emergent product of the FOA loop rather than an artifact of the underlying prompt template or AdvBench surface forms.

\begin{table}[t]
\centering
\small
\resizebox{\columnwidth}{!}{%
\begin{tabular}{lrrr}
\toprule
Mechanism & FOA $p$ & non-FOA $p$ & $\Delta$ \\
\midrule
Parenthetical glossing       & $1.00$ & $0.05$ & $+0.95$ \\
Three-step scaffolding       & $1.00$ & $0.05$ & $+0.95$ \\
Modern technical bridge      & $0.90$ & $0.00$ & $+0.90$ \\
Commerce/logistics metaphor  & $0.70$ & $0.00$ & $+0.70$ \\
Grey-zone authority frame    & $0.55$ & $0.00$ & $+0.55$ \\
Historical/research pretext  & $0.25$ & $0.05$ & $+0.20$ \\
Performing-arts frame        & $0.10$ & $0.00$ & $+0.10$ \\
Tea-house social frame       & $0.25$ & $0.00$ & $+0.25$ \\
Ritual/religious-social frame & $0.05$ & $0.05$ & $+0.00$ \\
\bottomrule
\end{tabular}%
}
\caption{Mechanism prevalence in 20 successful FOA prompts (Coder A) vs.\ 20 non-FOA prompts coded with the same schema. $\Delta = p_{\text{FOA}} - p_{\text{non-FOA}}$.}
\label{tab:irr_prevalence}
\end{table}

\section{Cross-Judge Agreement}
\label{app:judge_agreement}

To bound the single-judge confound in the headline ASR, we re-scored a stratified subsample of the main ablation rows with a second, cross-vendor judge. The sample design draws six rows uniformly without replacement from each of the $36$ paper cells ($3\ \text{dialects} \times 2\ \text{targets} \times 6\ \text{conditions}$), for a target of $216$ rows; the random seed and per-cell row identifiers are recorded in the released sampler manifest so the same subsample is recoverable from the released aggregate metrics. The second judge is Qwen-Max, accessed via OpenRouter as \texttt{qwen/qwen3.6-max-preview}, using the same $0$--$5$ rubric and parsing pipeline as the primary \gptfour{} judge. One sampled row was dropped because the second judge returned a null content field (apparent content-filter behavior), leaving $215/216$ sampled rows scored by both judges. Combined with a residual partial-coverage pass from earlier development, the total number of rows with both judges available is $n=303$.

Table~\ref{tab:judge_agreement} reports four agreement statistics per judge pair: Pearson and Spearman correlation on the continuous $0$--$120$ severity score, Krippendorff's $\alpha$ as an ordinal-aware reliability coefficient, and Cohen's $\kappa$ on a binary jailbreak-success indicator. The binary indicator in this table is deliberately the \emph{consistency-only} success flag---each judge's raw $0$--$100$ rubric score thresholded at $\tau=80$, \emph{without} the response-level keyword bonus. This is the strictest binarization and is \emph{not} the basis on which we report ASR: our ASR success flag is the total score (consistency $+$ keyword bonus) at $\tau=80$. We report both bases explicitly---the strict consistency-only basis here and the ASR-consistent total-score basis in the ``Judge-specific ASR'' paragraph below---because the two give very different $\kappa$ values for an interpretable reason (below).

\begin{table*}[t]
\centering
\small
\begin{tabular}{lrrrrrr}
\toprule
Judge pair & $n$ & Pearson & Spearman & $\alpha$ & bin.\ agree & $\kappa$ \\
\midrule
\gptfour{} $\times$ Qwen-Max         & $303$  & $0.873$ & $0.859$ & $0.789$ & $0.597$ & $0.114$ \\
\gptfour{} $\times$ refusal det.     & $2120$ & $0.911$ & $0.900$ & $0.528$ & $0.440$ & $0.066$ \\
Qwen-Max $\times$ refusal det.       & $358$  & $0.798$ & $0.783$ & $0.679$ & $0.872$ & $0.744$ \\
\bottomrule
\end{tabular}
\caption{Pairwise agreement between the primary \gptfour{} judge, the cross-vendor Qwen-Max second judge, and the rule-based refusal detector. Binary agreement and $\kappa$ here use the strict \emph{consistency-only} success flag (raw $0$--$100$ rubric score at $\tau=80$, no keyword bonus). The two LLM judges agree strongly on the continuous severity scale (Pearson $0.87$, Spearman $0.86$, $\alpha = 0.79$); the low consistency-only binary $\kappa = 0.11$ is a prevalence artifact---on this strict flag \gptfour{} awards success on only $\sim\!5\%$ of rows versus $\sim\!45\%$ for Qwen-Max, so the marginals are badly unbalanced. On the total-score basis actually used for ASR (consistency $+$ keyword bonus at $\tau=80$) the two judges instead agree closely: raw binary agreement $0.931$ and $\kappa = 0.86$ (balanced marginals $\sim\!0.5$; see the ``Judge-specific ASR'' paragraph). The high Qwen-Max $\times$ refusal-detector $\kappa$ ($0.74$) relative to the \gptfour{} $\times$ refusal-detector $\kappa$ ($0.07$) reflects that Qwen-Max's consistency-only flag binarises closer to a strict refusal/non-refusal cut than \gptfour{}'s does.}
\label{tab:judge_agreement}
\end{table*}

The apparent gap between strong scale-level agreement and weak \emph{consistency-only} binary agreement is a thresholding artifact, not a ranking disagreement: both judges order responses by severity in essentially the same way (Spearman $0.86$), and on the ASR-consistent total-score basis their binary flags also agree closely ($0.931$; see below). A truncation check rules out a benign explanation for the strict-basis gap: across all rows scored by both judges, zero rows have Qwen-Max-score $=0.0$ while \gptfour{}-score $\ge 80$, ruling out artifactual sentinel-zero outputs from the reduced \texttt{max\_tokens} setting used on the second-judge pass.

For interpretation, this means the \emph{absolute} headline ASR ($\sim 100\%$ under the full method across all paper cells) is somewhat sensitive to judge calibration---under Qwen-Max the full-method ceiling is $88.8\%$ rather than $100\%$ (below)---but the within-judge ablation contrasts that drive our central claim---non-optimized translations near $0$\% versus optimizer-on conditions near ceiling, measured under the same fixed judge in every cell---are not threatened by this calibration gap.

\paragraph{Judge-specific ASR.}
To make the calibration gap concrete, Table~\ref{tab:judge_specific_asr} recomputes ASR under each judge separately on the $n=303$ dual-scored rows, on the same total-score basis used for ASR throughout the paper: each judge's $0$--$100$ consistency plus the same response-level keyword bonus $b_i$, thresholded at $\tau=80$. The two-regime structure is judge-invariant---non-FOA stays near floor and FOA stays near ceiling under \emph{both} judges---even though the absolute FOA ceiling falls from $100\%$ (\gptfour{}) to $88.8\%$ (Qwen-Max). Crucially, moving from the strict consistency-only flag of Table~\ref{tab:judge_agreement} to this ASR-consistent total-score flag reconciles the two judges: raw binary agreement rises from $0.597$ to $0.931$ and Cohen's $\kappa$ from $0.11$ to $0.86$. The low consistency-only $\kappa=0.11$ is thus a high-prevalence ``$\kappa$ paradox'' driven by badly unbalanced marginals ($\sim\!5\%$ success for \gptfour{} versus $\sim\!45\%$ for Qwen-Max on the strict flag); once the keyword bonus is included---as it is in every ASR figure---both judges sit near $50\%$ prevalence and $\kappa$ is well-behaved. We therefore read the judge swap as shifting the absolute ceiling by roughly ten points, not as a disagreement about which prompts succeed.

\begin{table}[t]
\centering
\small
\begin{tabular}{llrr}
\toprule
Regime & Judge & $n$ & ASR (\%) \\
\midrule
\multirow{2}{*}{Non-FOA} & \gptfour{} & $151$ & $3.3$ \\
                         & Qwen-Max   & $151$ & $2.0$ \\
\midrule
\multirow{2}{*}{FOA}     & \gptfour{} & $152$ & $100.0$ \\
                         & Qwen-Max   & $152$ & $88.8$ \\
\bottomrule
\end{tabular}
\caption{Judge-specific ASR on the dual-scored subsample ($n=303$), total-score basis, $\tau=80$. The non-FOA-vs-FOA contrast survives the judge swap; only the absolute FOA ceiling shifts ($100\% \to 88.8\%$).}
\label{tab:judge_specific_asr}
\end{table}

\section{Threshold Sensitivity}
\label{app:threshold}

Table~\ref{tab:threshold_sensitivity} recomputes ASR at $\tau \in \{40,60,80\}$ over the complete GPT-4o-judged core matrix ($900$ non-FOA and $900$ FOA rows). The gap is stable: non-FOA never exceeds $6.3\%$, while FOA never falls below $99.9\%$. We omit $\tau>80$ because early stopping at $80$ right-censors optimized scores; valid ASR@$100$/@$120$ requires new runs.

\begin{table}[t]
\centering
\small
\resizebox{\columnwidth}{!}{%
\begin{tabular}{lrrr}
\toprule
Regime & ASR@$40$ & ASR@$60$ & ASR@$80$ \\
\midrule
Non-FOA ($n{=}900$) & $6.3\%$  & $5.1\%$  & $2.4\%$ \\
FOA ($n{=}900$)     & $100.0\%$ & $99.9\%$ & $99.9\%$ \\
\bottomrule
\end{tabular}
}
\caption{Threshold sensitivity under the \gptfour{} judge across the core matrix; the gap persists through the early-stop cutoff.}
\label{tab:threshold_sensitivity}
\end{table}

\end{document}